\documentclass[10pt,journal,cspaper,compsoc]{IEEEtran}
\usepackage{times}
\usepackage{epsfig}
\usepackage{graphicx}
\usepackage{amsmath}
\usepackage{amssymb}

\ifCLASSOPTIONcompsoc
\else
\fi
\ifCLASSINFOpdf
\else
\fi
\begin{document}
\title{Gaussian Affine Feature Detector}

\author{Xiaopeng~Xu,
        Xiaochun~Zhang
\IEEEcompsocitemizethanks{
\IEEEcompsocthanksitem X. Xu and X. Zhang are with Nanjing University of Science and Technology, China.}
\thanks{}}


\markboth{Journal of \LaTeX\ Class Files,~Vol.~6, No.~1, January~2007}%
{Shell \MakeLowercase{\textit{et al.}}: Bare Demo of IEEEtran.cls for Computer Society Journals}

%\IEEEpubid{\makebox[\columnwidth]{\hfill 0000--0000/00/\$00.00~\copyright~2007 IEEE}%

% PRIOR to the title within the \IEEEcompsoctitleabstractindextext IEEEtran
\IEEEcompsoctitleabstractindextext{%
\begin{abstract}
%\boldmath
A new method is proposed to get image features' geometric information.
Using Gaussian as an input signal, a theoretical optimal solution to calculate feature's affine shape is proposed. Based on analytic result of a feature model, the method is different from conventional iterative approaches. 
From the model, feature's parameters such as position, orientation, background luminance, contrast, area and aspect ratio can be extracted.

Tested with synthesized and benchmark data, the method achieves or outperforms existing approaches in term of accuracy, speed and stability. The method can detect small, long or thin objects precisely, and works well under general conditions, such as for low contrast, blurred or noisy images.
\end{abstract}
% not want either math or citations to appear in the abstract.

% Note that keywords are not normally used for peer review papers.
}

\maketitle

% is not selected <OR> if conference mode is selected - because compsoc
% conference papers position the abstract like regular (non-compsoc)
\IEEEdisplaynotcompsoctitleabstractindextext

% \fi
\IEEEpeerreviewmaketitle

\section{Introduction}
Detecting two dimension signals is more difficult than one dimension ones and many heuristic algorithms are proposed to deal with it. However, following appearance of scale space theory \cite{IEEEhowto:SS,IEEEhowto:koe84}, many effective feature detectors come into being \cite{IEEEhowto:Lin}. 

Originally, scale space theory is proposed by physicists, and developed by computer scientists. It is studied thoroughly from view point of vision and mathematics, and a consistent way to find new detector had been built \cite{IEEEhowto:Romeny}. Many successful feature detectors are built upon scale space, including	\cite{IEEEhowto:lowe04,IEEEhowto:lowe99}.

An additional dimension is introduced in scale space, namely scale dimension. 
In order to get image's information, such as affine shape parameters, some methods \cite{IEEEhowto:compare_feat,IEEEhowto:Eval_Feat,IEEEhowto:Mikolajczyk_IJCV} iteratively search in scale dimension. They are based on fix point theory: they will finally get a solution if there is one. In practice, however, these methods have several drawbacks, including,
  \begin{itemize}
    \item waste lots of candidate features;
    \item very slow;
    \item get abundant duplicated or false features.
  \end{itemize}

To overcome these drawbacks, we propose a new method based on analytic solution. It achieves or outperforms iterative methods  with much less computation resources.

Some feature detectors \cite{IEEEhowto:MSer,IEEEhowto:ibr_ebr} are ideal for noise free images, but are incapable of blurred or noisy images. The proposed detector is more robust with similar performance.

Features' position, orientation, background luminance, contrast, area and aspect ratio can also be extracted from images.
Until recently, information such as background luminance contrast are not commonly used in feature extraction. Others including area, orientation and aspect ratio are studied extensively, but with a limited accuracy.

In this paper, a feature model is proposed, and the above mentioned parameters will be calculated.
\section{Gaussian Affine Shape}
In this section, firstly a feature model is proposed, and then analytic result is derived based on the model to get various parameters.

\subsection{Feature Model}
From a view point of systematology, images, feature extractor and features correspond to input, system, and output. We need build a system that can transform input to output. In another words, image is system input and feature parameters are output. In order to study behavior of the system, we need define input signals.
As it is not possible to build an all-purpose feature extractor, we will concentrate on some specific image signals.
Since (two dimensional) Gaussian has nice analytic properties and simple form, it is chosen as input signal. As to be shown later, Gaussian based model will filter out high frequency signal, hence ideal for noisy images.

\begin{figure}% use float package if you want it here
  \centering
  \includegraphics[height=50mm]{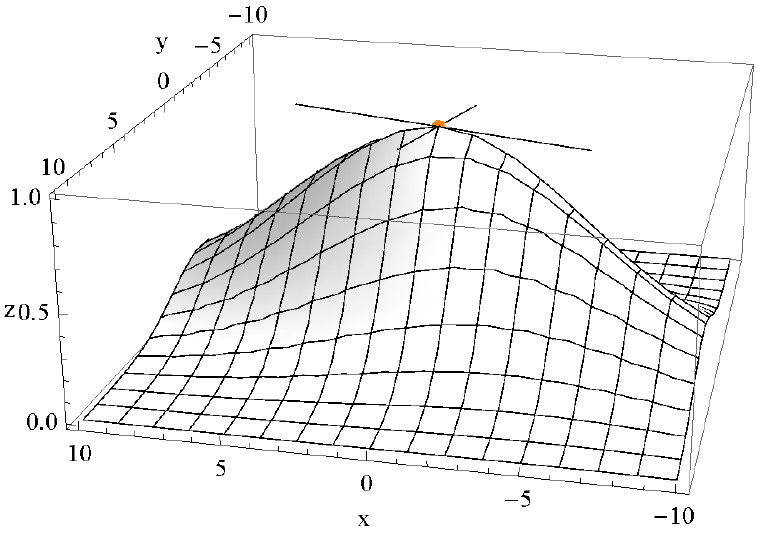}
  \caption{Model of input signal, whose baseline height is $0$ and contrast, is $1$.}
  \label{fig:gaussIn}
\end{figure}

Based on above mentioned idea, image surface is modeled as Gaussian function, as shown in Fig.~\ref{fig:gaussIn}. In this way, image feature parameters are related to Gaussian parameters, including orientation, long and short radii, baseline height and contrast. Traditionally,  baseline height and contrast are not considered in feature extraction, they are included for completeness. The signal can be defined as Equ.~\ref{equ:gaussIn}. Parameters and model variables are listed in Tab.~\ref{tab:model_factor}.

\begin{equation}
  \label{equ:gaussIn}
  f(x) = c\ e^{-\frac{1}{2}\left(\overset{\rightharpoonup }{x}-\overset{\rightharpoonup }{\mu }\right)^T\Sigma ^{-1}\left(\overset{\rightharpoonup }{x}-\overset{\rightharpoonup }{\mu }\right)}+d
\end{equation}

\begin{table}[!t]
\renewcommand{\arraystretch}{1.3}
\caption{Factors of model}
\label{tab:model_factor}
\centering
\begin{tabular}{c||c}
\hline
\bfseries Factors & \bfseries Parameters\\
\hline\hline
Contrast&$c$ \\
\hline
Baseline~height&$d$ \\
\hline
Long~radius&$\beta$ \\
\hline
Short~radius&$\alpha$ \\
\hline
Nominal~radius&$\sqrt{\alpha \beta}$\\
\hline
Aspect~ratio&$\frac{\beta}{\alpha}$ \\
\hline
Orientation&$\theta$ \\
\hline
LoG detected scale&$\sigma$ \\
\hline
\end{tabular}
\end{table}

Before continuing, it is helpful to clarify a fact, that is, rotating an image will not affect our discussion. This fact greatly simplifies our deduction. It is proved in Appendix.~\ref{proof}.

Known the fact, two-dimensional axis-aligned Gaussian will be used as input signal, as shown in Equ.~\ref{equ:sigin}. For this function, we need get value of $\alpha$, $\beta$, $c$ and $d$.
\begin{equation}
 \label{equ:sigin}
\text{I}\left(x,y;\alpha ,\beta ,c,d\right)=c\ e^{-\frac{1}{2}\left(\frac{x^2}{\alpha ^2}+\frac{y^2}{\beta ^2}\right)}+ d
\end{equation}

\subsection{Feature Detection}

Before computing parameters, we need detect feature's position. 
There exists many feature detectors, we need choose the one that has good performance and solid mathematical foundation. It will be chosen from rotation invariant differential operator family. As defined in Equ.~\ref{equ:LoG}, LoG detector is a good candidate, it is very stable, has fast implementation, and is Gaussian based. The last one is the most important reason, because input signal is also a Gaussian, and they may have close relation.
\begin{equation}
 \label{equ:LoG}
\text{LoG}= \nabla  \cdot  \nabla  G, G=\frac {e^{-\frac {x^2 + y^2} {2\sigma^2}}} {2\pi\sigma^2}
\end{equation}
\begin{eqnarray}
 \label{equ:convlrd}
 \partial _x\left(f_1*f_2\right) =f_1*\partial _xf_2\text\nonumber \\
 =\partial _xf_1*f_2
\end{eqnarray}

Using Equ.~\ref{equ:laplacelr}, which is based on Equ.~\ref{equ:convlrd}, we can define LoG operation on image function, as shown in Equ.~\ref{equ:lagconv}.
\begin{eqnarray}
 \label{equ:laplacelr}
{\partial _{x,x}}({f_1} * {f_2}) = {\partial _{x,x}}{f_1} * {f_2} = {f_1} * {\partial _{x,x}}{f_2}\nonumber\\
{\partial _{x,y}}({f_1} * {f_2}) = {\partial _{x,y}}{f_1} * {f_2} = {f_1} * {\partial _{x,y}}{f_2}\nonumber\\
{\partial _{y,y}}({f_1} * {f_2}) = {\partial _{y,y}}{f_1} * {f_2} = {f_1} * {\partial _{y,y}}{f_2}
\end{eqnarray}
\begin{eqnarray}
 \label{equ:lagconv}
\text{LoG} * I = \left(\partial _{x,x}G+\partial _{y,y}G\right)*I\nonumber\\
= \partial _{x,x}G*I+\partial _{y,y}G*I\nonumber\\
= \nabla  \cdot  \nabla (G*I)
\end{eqnarray}

As shown in Equ.~\ref{equ:convGI}, convolving I with G is another Gaussian, which is called (Gaussian) scale space. For zero shifted $I$, its Laplacian will get extreme at origin. Normalizing this value will get normalized Laplacian of Gaussian operation upon I, which is basis of some feature extractors.
\begin{eqnarray}
 \label{equ:convGI}
 G*I=\frac{c\alpha \beta  }{\sqrt{\sigma ^2+\alpha ^2} \sqrt{\sigma ^2+\beta ^2}}e^{-\frac{1}{2}\left(\frac{x^2}{ \left(\alpha ^2+\sigma ^2\right)}+\frac{y^2}{ \left(\beta ^2+\sigma ^2\right)}\right)}+ d
\end{eqnarray}

Applying LoG to image to get extreme points, and with information provided by $G*I$, we need get radii (standard deviations) of original input (Gaussian) function $I$.

\subsection{Parameters Calculation}

As shown before, image $I$ can be considered as a surface in three-dimensional space. From differential geometry, we know its hessian matrix directly relates to principal curvatures and principal directions, and for Gaussian function, principal curvatures connect with its standard deviations. In one word, eigenvalues of hessian matrix relate to radii and  eigenvectors relate to directions. We also know that two principal directions are perpendicular to one another. Based on these facts, we will derive formulas for parameters.

Obviously, convolving $I$ with isotropic Gaussian will not change principal direction. For extreme point, we can use $G*I$'s principal direction as $I$'s principal direction. For the case of axis-aligned Gaussian, we already know principal directions, otherwise, compute eigenvectors.

Remaining question is, giving information of $G*I$ , how to get $I$'s radii $\alpha$ and $\beta$, its contrast $c$ and baseline height $d$.

Here, we will exploit a fact, that LoG can detect Gaussian at one and only one scale. In another word, every $\alpha$ and $\beta$ pair must produce one and only one $\sigma$, as shown in Equ.~\ref{equ:constrain}. If analytic form of $f$ is determined, we can recover $\alpha$ and $\beta$ from $\sigma$.

\begin{eqnarray}
 \label{equ:constrain}
\sigma =f(\alpha ,\beta )
\end{eqnarray}

For any input image, $\frac{\beta }{\alpha }$ is fixed. LoG will detect extreme point in a fixed scale $\sigma$. Let us denote $k=\frac{\beta }{\alpha }$, and $h=\frac{\alpha }{\sigma }$.

Apply normalized LoG operator to I, and substitute $\beta  = \alpha  k$ and $\alpha  = \sigma  h$, and let $x=0$, $y=0$, we get Equ.~\ref{equ:LoGMax}.

\begin{equation}
 \label{equ:LoGMax} \sigma ^2\nabla  \cdot  \nabla  (G*I)_{x=0,y=0}=-\frac{c h^2 k \left(2+h^2 \left(1+k^2\right)\right)}{\left(\left(1+h^2\right) \left(1+h^2 k^2\right)\right)^{3/2}}
\end{equation}

\begin{figure}% use float package if you want it here
  \centering
  \includegraphics[height=40mm]{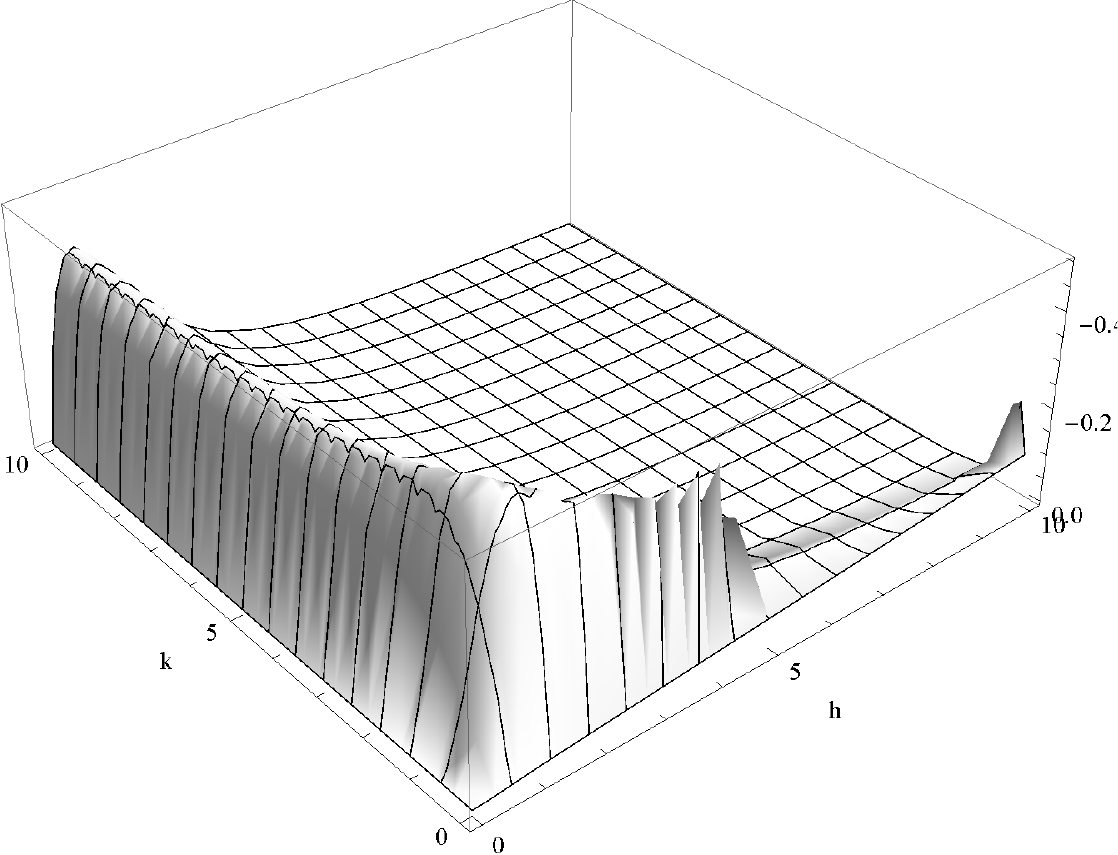}
  \caption{k and h constrained}
  \label{fig:ridge}
\end{figure}

Let $c$ be constance $1$ and draw this expression in Fig.~\ref{fig:ridge}. It is clearly shown that for $k>1$, extreme of $LoG*I$ is located on a smooth ridge.

For a fixed $k$, at extreme point, the formula's one order derivative will be zero. After some calculation, we can get  Equ.~\ref{equ:LoGMaxDeriv}.

\begin{equation}
 \label{equ:LoGMaxDeriv}
-4-2 h^2 \left(1+k^2\right)-h^4 \left(1-6 k^2+k^4\right)+2 h^6 \left(k^2+k^4\right)==0
\end{equation}

To solve this equation, let $k^2=K$ and $h^2=H$, and we get two order equation Equ.~\ref{equ:twoorderequ}.
\begin{equation}
 \label{equ:twoorderequ}
 -4-2 H-H^2+\left(-2 H+6 H^2+2 H^3\right) K+\left(-H^2+2 H^3\right) K^2
\end{equation}

It is easy to solve, as Equ.~\ref{equ:sol}.
\begin{eqnarray}
 \label{equ:sol}
K_1=\frac{1-3 H-H^2-(1+H) \sqrt{-3+H (6+H)}}{H (-1+2 H)} \nonumber\\
K_2=\frac{1-3 H-H^2+(1+H) \sqrt{-3+H (6+H)}}{H (-1+2 H)}
\end{eqnarray}

Known constraint of $K$ and $H$, we need more information to get their values.
As mentioned above, eigenvalues of hessian matrix relate to radii closely. We calculate hessian matrix over scale space, as shown in Equ.~\ref{equ:hessscale}.

\begin{equation}
 \label{equ:hessscale}
\left(
\begin{array}{cc}
 \partial _{x,x} & \partial _{x,y} \\
 \partial _{x,y} & \partial _{y,y}
\end{array}
\right)(G*I)
\end{equation}

As before, we calculate eigenvalues of this matrix, and let $x=0$, $y=0$. Since our discuss based Gaussian, we can get analytic solution of two eigenvalues, shown in Equ.~\ref{equ:eig}.

\begin{eqnarray}
 \label{equ:eig}
\text{e}_1=-\frac{c h^2 k}{\left(1+h^2\right)^{3/2} \sqrt{1+h^2 k^2} \sigma ^2} \nonumber\\
\text{e}_2=-\frac{c h^2 k}{\sqrt{1+h^2} \left(1+h^2 k^2\right)^{3/2} \sigma ^2}
\end{eqnarray}

These two eigenvalues have complicated form, but their ratio is simpler, shown in Equ.~\ref{equ:eigratio}.

\begin{equation}
 \label{equ:eigratio}
 r=\frac{\text{e}_1}{\text{e}_2}=\frac{1+H K}{1+H}
\end{equation}

From Equ.~\ref{equ:eigratio}, we can solve for $K$, shown in Equ.~\ref{equ:KR}.
\begin{equation}
 \label{equ:KR}
K=\frac{-1+r+H r}{H}
\end{equation}

Combined Equ.~\ref{equ:sol} and Equ.~\ref{equ:KR}, we can solve for $H$, result is Equ.~\ref{equ:Hres}.
\begin{equation}
 \label{equ:Hres}
H= \frac{3+r^2}{2 r (1+r)}
\end{equation}

\begin{figure}% use float package if you want it here
  \centering
  \includegraphics[height=40mm]{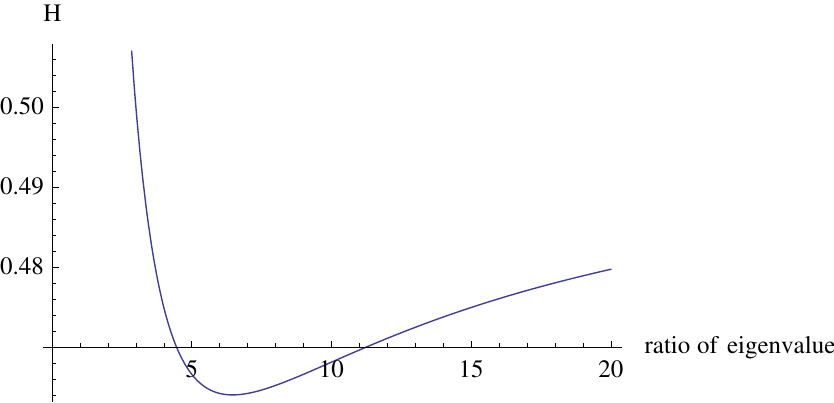}
  \caption{Relation of H and ratio of eigenvalues}
  \label{fig:Hr}
\end{figure}

We draw this relation in Fig.~\ref{fig:Hr}, which shows detecting scale tends to be constancy as shape gets elongate. Simply put, elongating a shape contributes little to its detecting scale.

Got $H$, $k$, $h$, $\alpha$ and $\beta$ will be solved directly.

$c$ and $d$ can also be solved in analytic form. Equ.~\ref{equ:LoGMax} is used to get $c$.
Because $d$ is constant component of scale space, it will disappear by  differential operation; therefore can only be solved in scale space itself. Let $x=0$ and $y=0$ in $G*I$, we will get Equ.~\ref{equ:scalemax}, so $d$ can be solved upon extreme point of scale space.
\begin{equation}
 \label{equ:scalemax}
G*I_{x=0,y=0}=d+\frac{c}{\sqrt{1+\frac{1}{H}} \sqrt{1+\frac{1}{H K}}}
\end{equation}

Until now, we have calculated all parameters of the zero shifted and axis aligned Gaussian.  Because axis can be shifted or rotated, our discussion will be applied to Gaussian of any position or rotation.
We will summary the steps of our algorithm.
  \begin{itemize}
    \item Detect extreme point in normalized LoG space, and get its $\sigma$.
    \item compute hessian matrix of extreme point in corresponding scale space
    \item compute eigenvectors as principal directions of the point.
    \item compute eigenvalues, let absolute larger one divide smaller one, and represented as $r$
    \item use Equ.~\ref{equ:Hres} to compute $H$,  Equ.~\ref{equ:KR} to compute K, and use $ \alpha =\sqrt{H} \sigma ,\beta = \sqrt{K} \alpha $ to compute other parameters, use Equ.~\ref{equ:LoGMax} and Equ.~\ref{equ:scalemax} to solve for contrast and baseline height.
  \end{itemize}

\subsection{Data Transformation}
Until now, signal's radii and angle are extracted. In order to comparing with other methods' results, we depend on some publicly available tools. Therefore radii and angle need to be transformed to a common form, such as symmetric positive definite matrix, as shown in Equ.~\ref{equ:symmat}.

\begin{equation}
 \label{equ:symmat}
\left( {\begin{array}{*{20}{c}}
x&y\\
y&z
\end{array}} \right)
\end{equation}

If let $\theta$ be signal's orientation, and $t = \arctan (\theta )$, in a similar way as before, we get Equ.~\ref{equ:xyzs}.

\begin{equation}
 \label{equ:xyzs}
\begin{array}{l}
x = \frac{{\beta  + {t^2}\alpha }}{{1 + {t^2}}}\\
y = \frac{{t(\alpha  - \beta )}}{{1 + {t^2}}}\\
z = \frac{{\alpha  + {t^2}\beta }}{{1 + {t^2}}}
\end{array}
\end{equation}

\section{Implementation Details}

In this section, some important implementation details are outlined. 
%Andrea Vedaldi opens an implementation of SIFT, which including scale space and DoG(LoG).
\subsection{Approximation and Adjustment}

As shown in  Equ.~\ref{equ:lowequ}, LoG can be implemented by DoG, and together with pyramid algorithm, which makes proposed method ready for application.
We use similar DoG pyramid as Lowe's.
Extremum of DoG should be adjusted by a constant multiplier, for its value is used to compute $c$ and $d$. 
\begin{equation}
 \label{equ:lowequ}
G(x,y,k \sigma )-G(x,y,\sigma )\approx (k-1)\sigma ^2 \nabla ^2G
\end{equation}
\subsection{Removal of False Features}

Tested with synthesized data, we found one common problem among several (affine) feature detectors, that is, for a single Gaussian signal, often there are several features detected out. Some of them have similar radii and orientations, located around true position, as shown in Fig.~\ref{fig:compare_circle} and Fig.~\ref{fig:noise_localvar}. Others are false features arisen from noise, as shown in Fig.~\ref{fig:noise_localvar} and Fig.~\ref{fig:compare_dist_noisy}. 
%There are two methods to amend the problem, and will to be disscussed in turn. Many false features can be removed in this way.
%Previously, it is assumed that feature located in extreme point of scale space, however extremum of LoG is used to detect feature points. Therefor only extreme point of scale space should be retained.  Based on similiar idea, Lowe et. al. refine SIFT features' position, but they do not remove unqualified features like us.

In practice, we found a large part of false features coming from sampling and digitization process, that is to say, they are small sized, low contrast features. True features seldom have such properties. Therefore features with small value of $c$, $\alpha$ and $\beta$ are considered as noises. 

\subsection{Detector Threshold}

Like SIFT, we uses ratio of principal curvatures (ratio of hessian's eigenvalues, or $r$ in our method) to remove points on valley or ridge. To accept more features, the ratio needs to be refined.
Combining Equ.~\ref{equ:Hres} and Equ.~\ref{equ:KR}, with $K = k^2$, we have Equ.~\ref{equ:kr-relation}.
\begin{equation}
 \label{equ:kr-relation}
k=\sqrt{\frac{r+3 r^3}{3+r^2}}
\end{equation}

\begin{figure}% use float package if you want it here
  \centering
  \includegraphics[height=40mm]{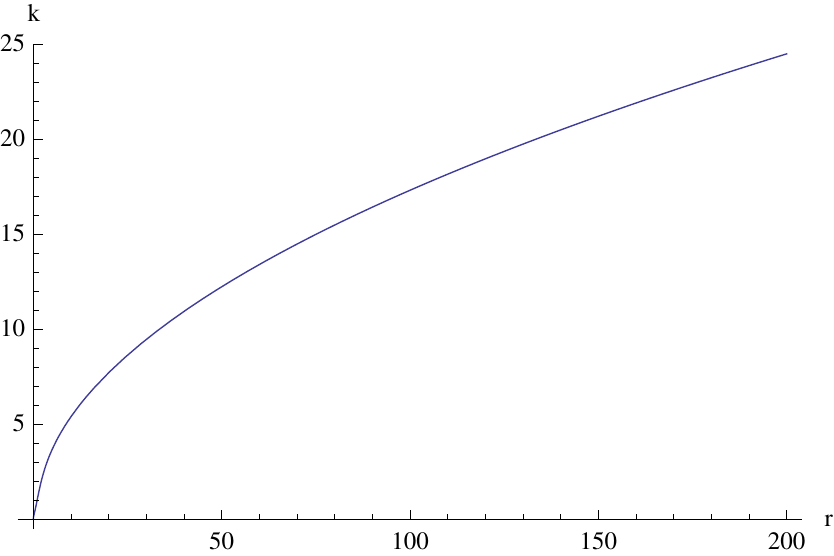}
  \caption{relation of $k$ and $r$}
  \label{fig:kr-relation}
\end{figure}

We have drawn relation of $k$ and $r$ in Fig.~\ref{fig:kr-relation}. For aspect ratio $k$ to be as high as $40$, $r$ need at least to be $535$ theoretically. The $r$ in Equ.~\ref{equ:throsh} is threshold of features.
\begin{equation}
 \label{equ:throsh}
\frac{\text{Tr}(H)^2}{\text{Det}(H)}=\frac{\left(e_1+e_2\right){}^2}{e_1e_2}=\frac{\left(r e_2+e_2\right)}{r e_2{}^2}=\frac{(r+1)^2}{r}
\end{equation}

\section{Experiments}

In order to evaluate performance of our method, we firstly test it with synthesized data. In this way, we will know true parameters and therefor can compare them with calculated ones. We will compare results of our method and others, including Harris-Affine, Hessian-Affine and Mser.
Only common parameters such as orientation, long and short scale can be compared, because contrast and base height are unique provided by our method. Nevertheless, we will show the results alone.

Gaussian will be used as test image. Image size is 256x256, and gray scale level is $256$. Our method can detect a large range of parameters, and Tab.~\ref{tab:param} lists parameters used in experiments.

\begin{table}[!t]
\renewcommand{\arraystretch}{1.3}
\caption{Test Image condition}
\label{tab:param}
\centering
\begin{tabular}{c||c}
\hline
\bfseries Parameter & \bfseries Range\\
\hline\hline
$c$ & $\left[ {-255,255} \right]$\\
\hline
$d$ & $\left[ {0,255} \right]$\\
\hline
$\sqrt{\alpha \beta}$ & $\left[ {5,40} \right]$\\
\hline
$\frac{\beta }{\alpha}$ & $\left[ {1,30} \right]$\\
\hline
$\theta$ & $\left[ { - \frac{\pi }{2},\frac{\pi }{2}} \right]$ \\
\hline
\end{tabular}
\end{table}
\subsection{Results of ideal signals}

\begin{figure}[t]
  \begin{center}
    %\fbox{\rule{0pt}{2in} \rule{0.9\linewidth}{0pt}} left bottom right   top
    \includegraphics[width= 0.5in, trim = 3.5in 5in 4in 4in clip]{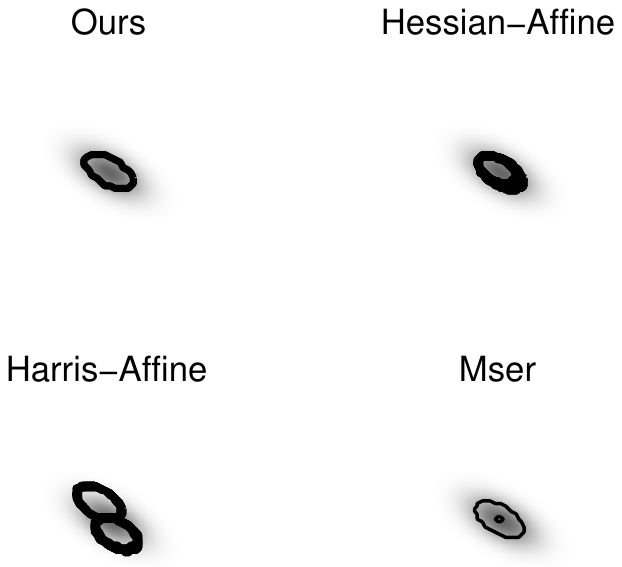}
  \end{center}
  \caption{Typical results for an isotropic Gaussian}
  \label{fig:compare_circle}
\end{figure}

\begin{figure}[t]
  \begin{center}
    %\fbox{\rule{0pt}{2in} \rule{0.9\linewidth}{0pt}} left bottom right   top
    \includegraphics[width= 0.38in, trim = 3.5in 4in 4in 3.5in clip]{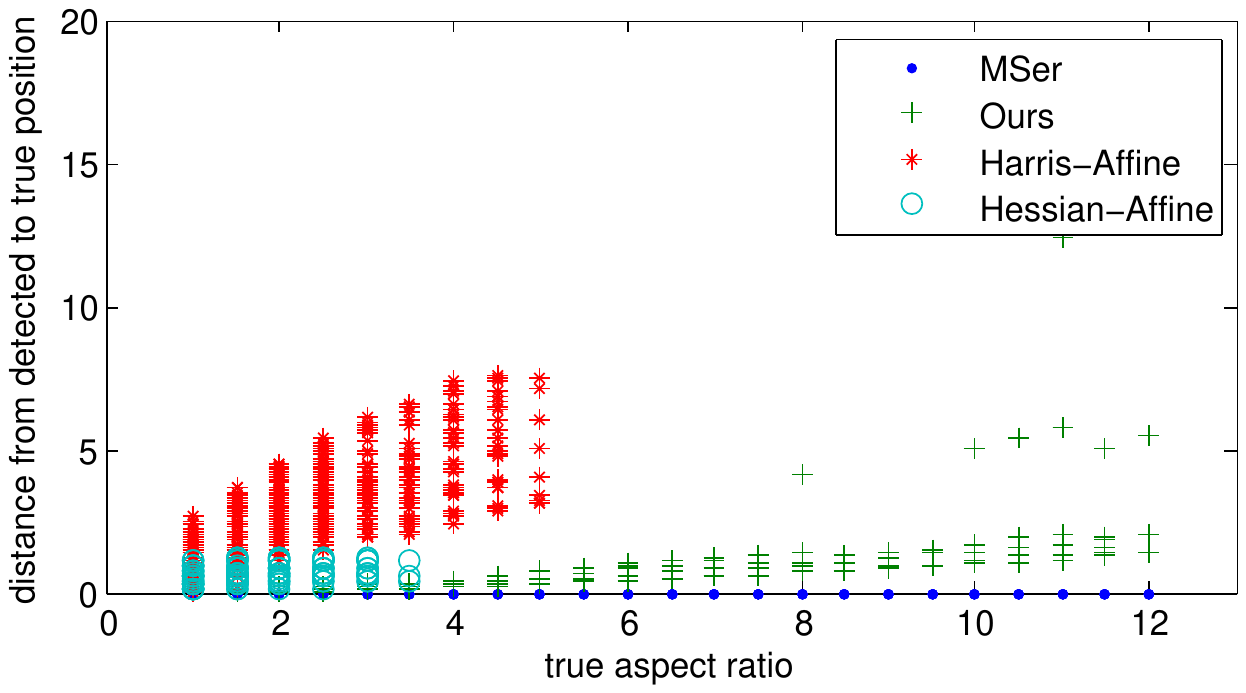}
  \end{center}
  \caption{Position inaccuracy}
  \label{fig:compare_dist}
\end{figure}

\begin{figure}[t]
  \begin{center}
    %\fbox{\rule{0pt}{2in} \rule{0.9\linewidth}{0pt}} left bottom right   top
    \includegraphics[width= 0.38in, trim = 3.5in 4in 4in 4.0in clip]{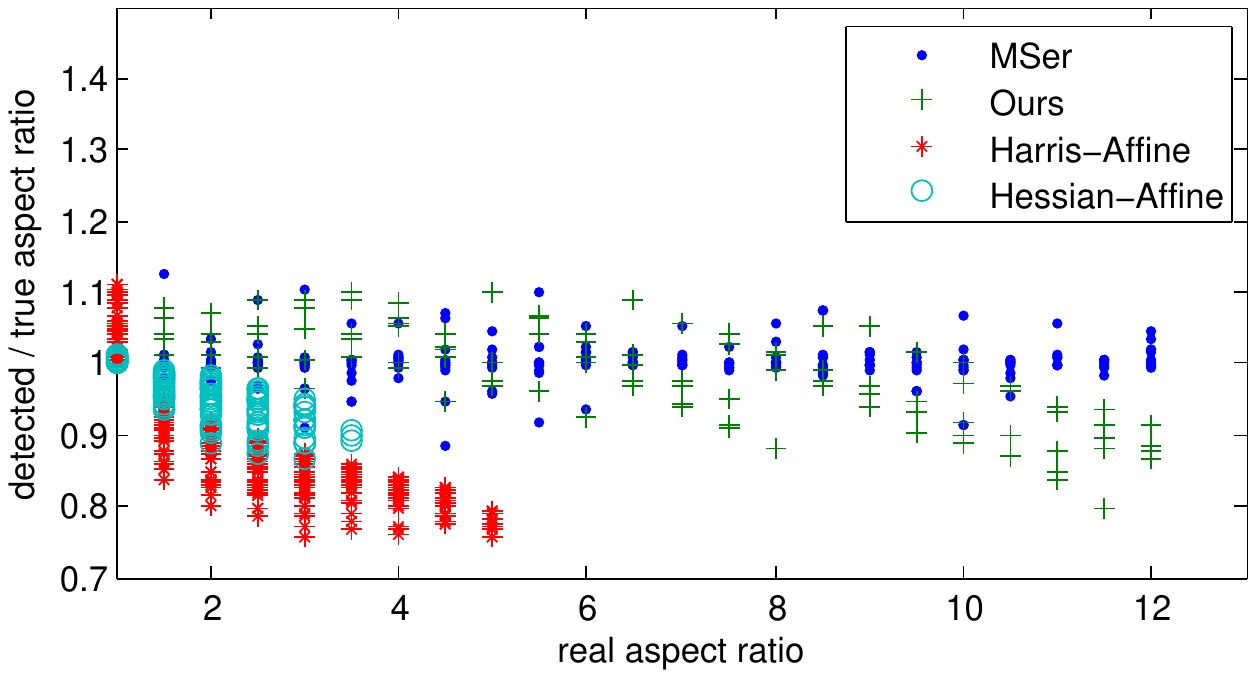}
  \end{center}
  \caption{Aspect ratio accuracy}
  \label{fig:aspect_ratio_nonoise}
\end{figure}

As demostrated in Fig.~\ref{fig:compare_circle}, Hessian-Affine and Harris-Affine tend to detect duplicated features. Fig.~\ref{fig:compare_dist} and Fig.~\ref{fig:aspect_ratio_nonoise} show, for noise free Gaussian signal, Mser has highest accuracy for detecting position and aspect ratio. Our method achieves similar results as MSer. Compared with Harris-Affine, Hessian-Affine gets better results. Both Mser and our method can detect signals of high aspect ratio, but Hessian-Affine and Harris-Affine are limited to low aspect ratio signals.

Our method is to compute original parameters from blurred output image. For very long and thin shapes, our method may slightly underestimate true aspect ratio, as shown in Fig~.\ref{fig:aspect_ratio_nonoise}. 

\begin{figure}[t]
  \begin{center}
    %\fbox{\rule{0pt}{2in} \rule{0.9\linewidth}{0pt}} left bottom right   top
    \includegraphics[width= 0.38in, trim = 3.5in 3.5in 4in 3.5in clip]{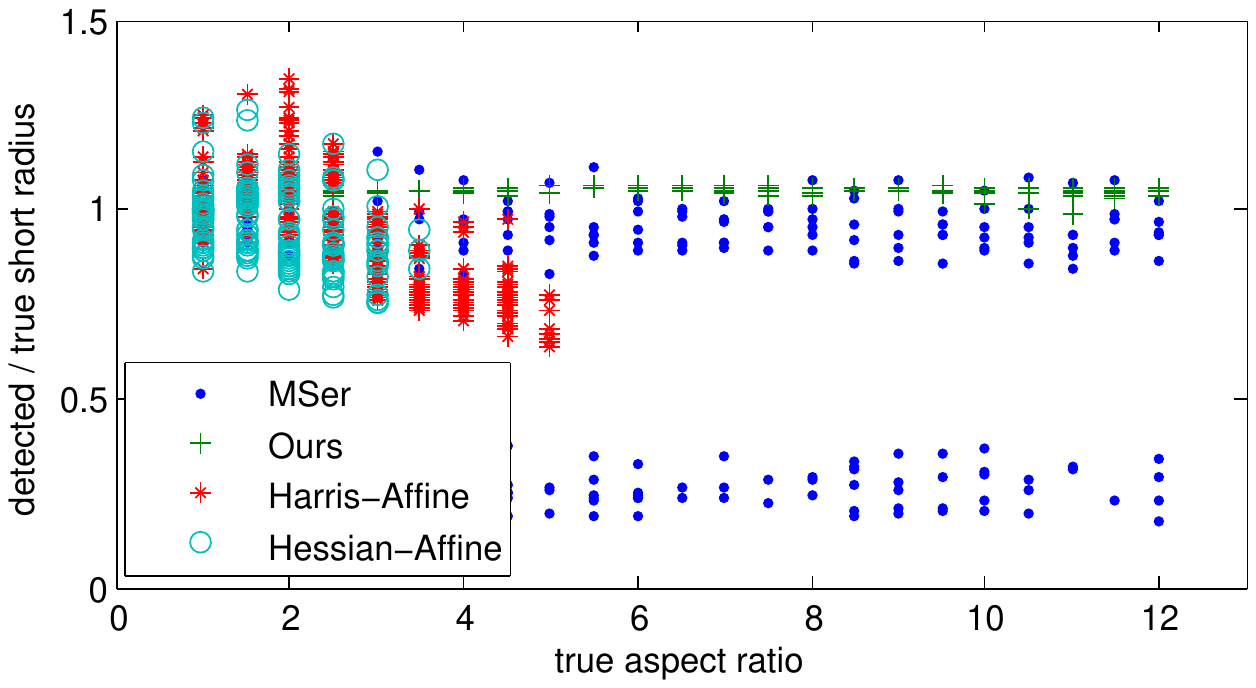}
  \end{center}
  \caption{Short radius accuracy}
  \label{fig:compare_short_scale}
\end{figure}

As shown in Fig.~\ref{fig:compare_short_scale}, our method and Mser achieve highest accuracy for detecting short radii. However, in addition to true signals, Mser often finds small concentric signals.

In conclusion, for ideal noise free Gaussian, Mser get best results, and ours is similar to that of Mser. Hessian-Affine and Harris-Affine are not as stable as Mser and ours.
\subsection{Results of noisy signals}
\begin{figure}[t]
  \begin{center}
    %\fbox{\rule{0pt}{2in} \rule{0.9\linewidth}{0pt}} left bottom right   top
    \includegraphics[width= 0.45in, trim = 3.5in 4in 4in 3.5in clip]{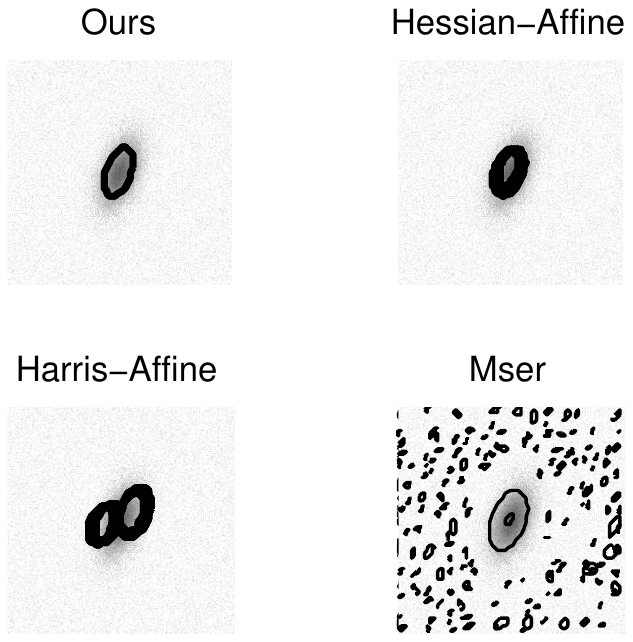}
  \end{center}
  \caption{Typical results for noisy Gaussian}
  \label{fig:noise_localvar}
\end{figure}
\begin{figure}[t]
  \begin{center}
    %\fbox{\rule{0pt}{2in} \rule{0.9\linewidth}{0pt}} left bottom right   top
    \includegraphics[width= 0.38in, trim = 3.5in 4in 4in 3.5in clip]{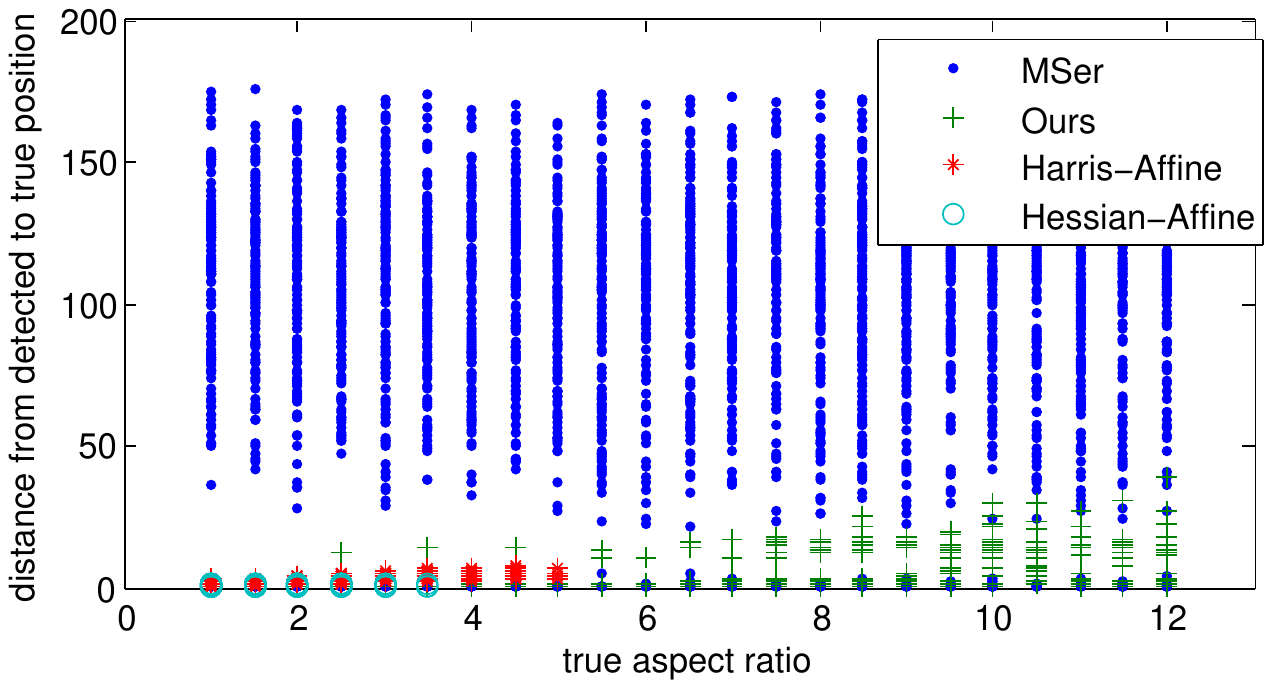}
  \end{center}
  \caption{Position inaccuracy for noisy Gaussian}
  \label{fig:compare_dist_noisy}
\end{figure}
\begin{figure}[t]
  \begin{center}
    %\fbox{\rule{0pt}{2in} \rule{0.9\linewidth}{0pt}} left bottom right   top
    \includegraphics[width= 0.38in, trim = 3.5in 4in 4in 3.5in clip]{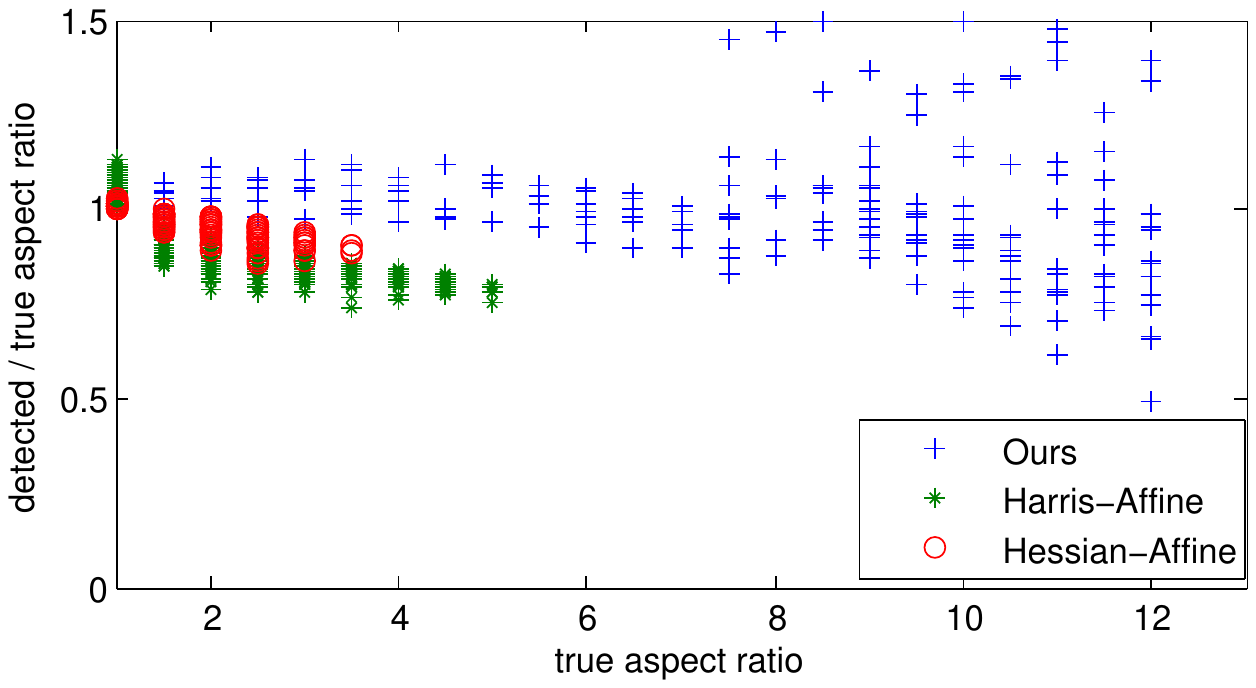}
  \end{center}
  \caption{Aspect ratio accuracy for noisy Gaussian}
  \label{fig:aspect_ratio_noisy}
\end{figure}

\begin{figure}[t]
  \begin{center}
    %\fbox{\rule{0pt}{2in} \rule{0.9\linewidth}{0pt}} left bottom right   top
    \includegraphics[width= 0.38in, trim = 3.5in 3.5in 4in 3.5in clip]{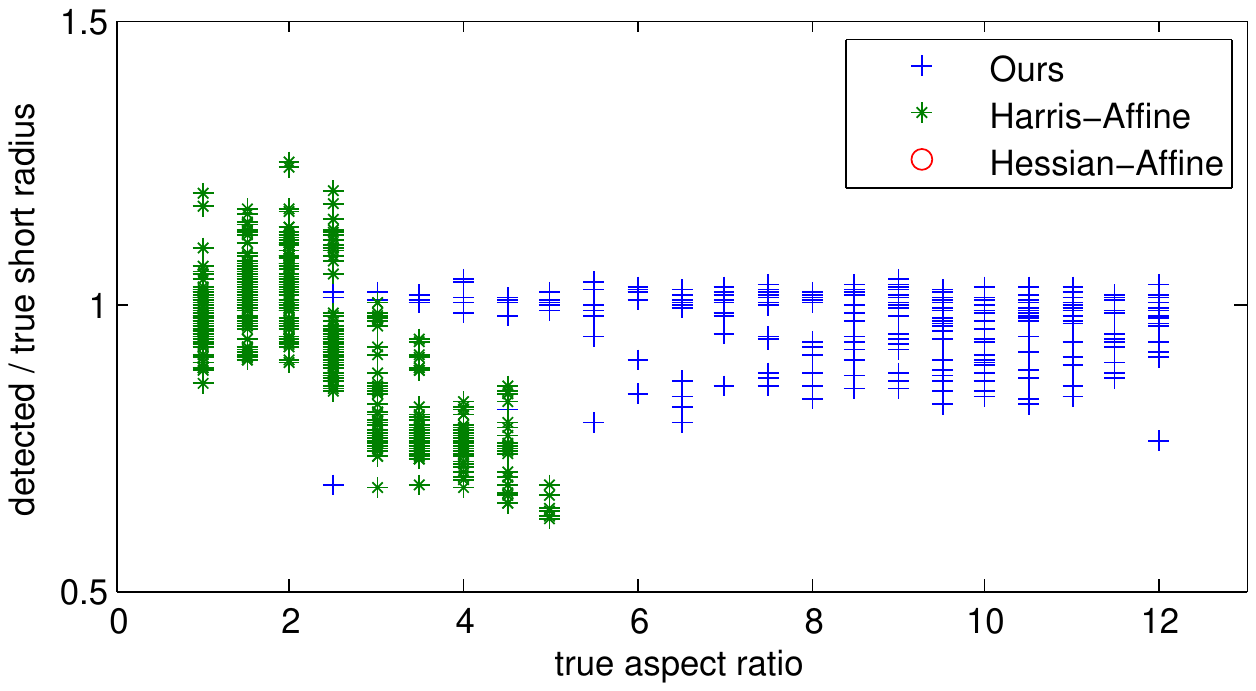}
  \end{center}
  \caption{Short radius accuracy for noisy Gaussian}
  \label{fig:compare_short_scale_noisy}
\end{figure}

\begin{figure}[t]
  \begin{center}
    %\fbox{\rule{0pt}{2in} \rule{0.9\linewidth}{0pt}} left bottom right   top
    \includegraphics[width= 0.38in, trim = 3.5in 3.5in 4in 3.5in clip]{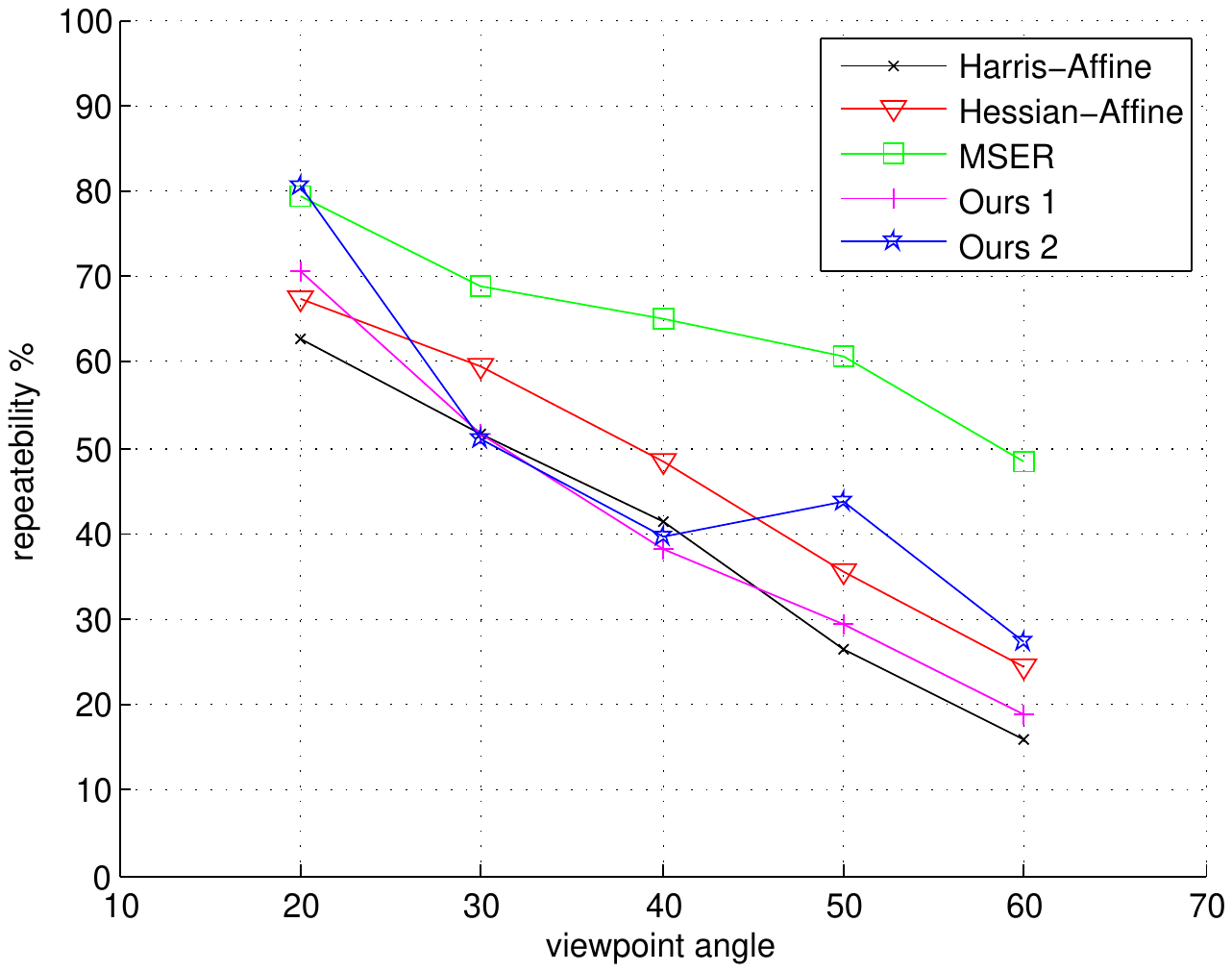}
  \end{center}
  \caption{Repeatebility comparasion. Our $1$ and $2$ have different threshold.}
  \label{fig:rep}
\end{figure}

\begin{figure}[t]
  \begin{center}
    %\fbox{\rule{0pt}{2in} \rule{0.9\linewidth}{0pt}} left bottom right   top
    \includegraphics[width= 0.55in, trim = 3.5in 2.5in 4in 2.5in clip]{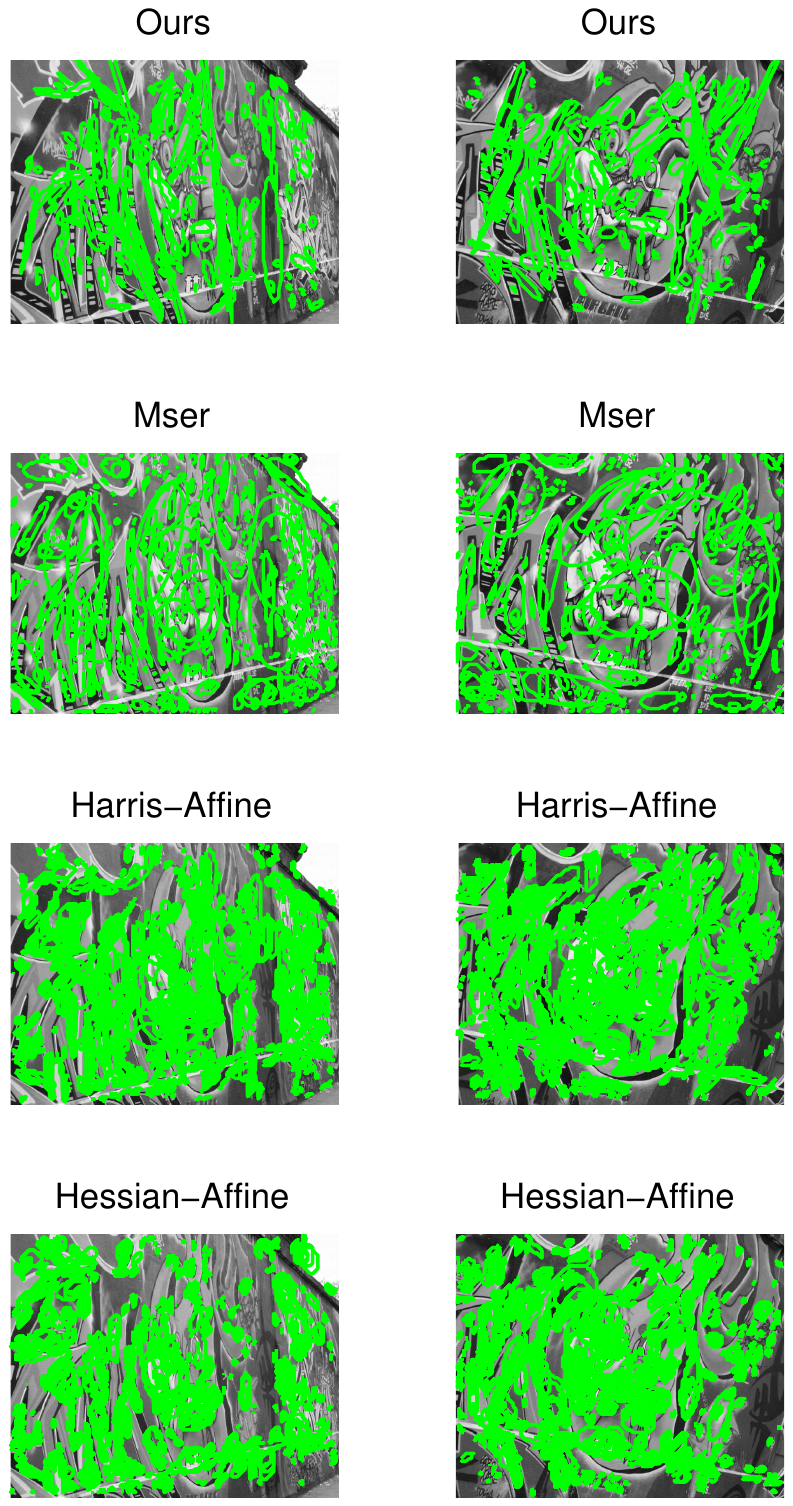}
  \end{center}
  \caption{Graffiti of different view angles.}
  \label{fig:compare_wall}
\end{figure}

Fig.~\ref{fig:noise_localvar} is a typical noisy image, and Mser is the most sensitive to noise. Even a small amount of noise can impact Mser seriously. Fig.~\ref{fig:compare_dist_noisy} is distance of true and detected points. It is difficult for Mser to differentiate noises from true signals. Therefore we only compare Hessian-Affine, Harris-Affine and ours for noisy images.

As shown in Fig.~\ref{fig:compare_dist_noisy}, Fig.~\ref{fig:aspect_ratio_noisy}, Fig.~\ref{fig:aspect_ratio_noisy}, our method performs well when other methods reach their limits.

Using Mikolajczyk's evaluation images and toolbox, we get repeatability in Fig.~\ref{fig:rep}.  For these noisy free images, Mser get highest accuracy, and Hessian-Affine, Harris-Affine and our methods have similar results. Our $1$ and $2$ are results of different thresholds.

Fig.~\ref{fig:compare_wall} is detecting results of graffiti under different view angle. Compared with Hessian-Affine and Harris-Affine, Mser and ours detect fewer features. It seems that the former two detect many redundant features. Compared with ours, Mser tends to detect many small features.
\section{Conclusion}

In this paper, we have proposed a new feature detector. Compared with other methods, it is very stable, accurate and quick. Tested with Gaussian, for ideal noisy free signal, our method produces one of the best results, and for noisy signal, it outperforms others significantly.
The proposed method can also extracts parameters unavailable for other methods, such as contrast and baseline height.

Test with benchmark images, the method get similar repeatability as Harris-Affine and Hessian-Affine.
%\appendix[Proof of Rotation Invariant for Image Surface]
% or


\appendices
\section{Proof of Rotation Invariant for Image Surface}
\label{proof}
Let $F$ be Fourier operator, and $f$ be an input function; Fourier transform is shown in Equ.~\ref{equ:fourtran}.
\begin{equation}
  \label{equ:fourtran}
  F\circ f(x)=\int _{\infty }^{\infty }f(x)e^{-2\pi  i\langle x, \xi \rangle }dx=\overset{\wedge }{f}(\xi )
\end{equation}
If input function rotates in $x$ space, and let \text{$\xi = R  x$}, its Fourier transform will also rotate same angle, as shown in Equ.~\ref{equ:fourtranrot}.
\begin{eqnarray}
\label{equ:fourtranrot}
F\circ f(R x)=\int _{\infty }^{\infty }f(R x)e^{-2\pi  i\langle x, \xi \rangle }dx\nonumber \\
=\pmb{ }\int _{\infty }^{\infty }f(y)e^{-2\pi  i\left\langle R^Ty, \xi \right\rangle }dy \nonumber \\
=\int _{\infty }^{\infty }f( y)e^{\left.-2\pi  i (R^Ty \right)^T\xi }dy \nonumber \\
=\int _{\infty }^{\infty }f( y)e^{-2\pi  iy ^TR \xi }dy \nonumber \\
=\int _{\infty }^{\infty }f(y)e^{-2\pi  i\langle y, R \xi \rangle }dy\text{  }\nonumber \\
= \overset{\wedge }{f}(R \xi )
\end{eqnarray}

Convolution in space domain can be implemented by multiplication in $\xi$ domain, as shown in Equ.~\ref{equ:convfour}.

\begin{equation}
 \label{equ:convfour}
 f(x)*g(x)\leftrightarrow \overset{\wedge }{f}(\xi )\overset{\wedge }{g}(\xi )=\overset{\wedge }{h}(\xi )\leftrightarrow h(x)
\end{equation}

\begin{equation}
 \label{equ:convfourrot}
f (R x) * g (R x) \leftrightarrow  \overset{\wedge }{f} (R \xi ) \overset{\wedge }{g} (R \xi ) = \overset{\wedge }{h} (R \xi ) \leftrightarrow  h (R x)
\end{equation}
Using Equ.~\ref{equ:fourtran}, Equ.~\ref{equ:fourtranrot} and Equ.~\ref{equ:convfour}, we can get Equ.~\ref{equ:convfourrot}. It means if input and system are rotated with same angle, the output will also rotate the same angle. In one word, output's geometrical property will not change on rotating input and system.

%\section{}

\ifCLASSOPTIONcompsoc
  \section*{Acknowledgments}
\else
  \section*{Acknowledgment}
\fi

The authors would like to thank PhD. Andrea Vedaldi for his excellent open sourced code, and professor Bart ter Haar Romeny for his free distributed electronic book.

\ifCLASSOPTIONcaptionsoff
  \newpage
\fi

\end{document}